\documentclass[letterpaper, 10 pt, conference]{ieeeconf} 

\IEEEoverridecommandlockouts                              

\usepackage{graphics} 
\usepackage{epsfig} 
\usepackage{mathptmx} 
\usepackage{times} 
\usepackage{amsmath} 
\usepackage{amssymb}  
\usepackage{mathtools}
\DeclarePairedDelimiter{\abs}{\lvert}{\rvert}
\usepackage{amsmath} 
\newcommand{\norm}[1]{\left\lVert#1\right\rVert}
\usepackage{mathtools,amssymb}
\usepackage{url}

\title{\LARGE \bf
A Whole-Body Model Predictive Control Scheme Including External Contact Forces and CoM Height Variations
}

\author{Reihaneh Mirjalili$^{1}$, Aghil Yousefi-koma$^{1}$, Farzad A. Shirazi$^{2}$, Arman Nikkhah$^{1}$, \\
Fatemeh Nazemi$^{1}$ and Majid Khadiv$^{3}$
\thanks{$^{1}$Center of Advanced Systems and Technologies (CAST), School of Mechanical Engineering, College of Engineering, University of Tehran, Tehran, Iran.
        {\tt\small aykoma@ut.ac.ir}}%
\thanks{$^{2}$School of Mechanical Engineering, College of Engineering, University of Tehran, Tehran, Iran.}%
\thanks{$^{3}$Movement Generation and Control group, Max-Planck Institute for Intelligent Systems, Tuebingen, Germany.}%
}

\begin{document}

\maketitle
\thispagestyle{empty}
\pagestyle{empty}

\begin{abstract}

In this paper, we present an approach for generating a variety of whole-body motions for a humanoid robot. We extend the available Model Predictive Control (MPC) approaches for walking on flat terrain to plan for both vertical motion of the Center of Mass (CoM) and external contact forces consistent with a given task. The optimization problem is comprised of three stages, i. e. the CoM vertical motion, joint angles and contact forces planning. The choice of external contact (e. g. hand contact with the object or environment) among all available locations and the appropriate time to reach and maintain a contact are all computed automatically within the algorithm. The presented algorithm benefits from the simplicity of the Linear Inverted Pendulum Model (LIPM), while it overcomes the common limitations of this model and enables us to generate a variety of whole body motions through external contacts. Simulation and experimental implementation of several whole body actions in multi-contact scenarios on a humanoid robot show the capability of the proposed algorithm. 
\end{abstract}

\section{Introduction}

The DARPA Robotics Challenge (DRC) \cite{darpa2015drc} showcased the state of the art capability of humanoid robots to perform some preliminary actions required for disaster relief scenarios. During the challenge, many robots lost their balance and fell down, while they could have grabbed the environment by their hands to prevent falling. Although most of the falls were due to hardware issues, the challenge revealed the lack of theory in synthesizing reactive planners to establish new contacts (especially hand contact) in response to unknown disturbances. we have not seen any robot during the challenge to even try to use hand contact to prevent falling down.

Establishing and removing contacts (switching dynamics) and controlling contact forces (continuous dynamics) are the main aspects of legged robots motion generation. Many approaches treat the problem of trajectory optimization by generating a sequence of contacts, and then optimizing the contact forces using this contact set \cite{dai2016planning,carpentier2017multi,ponton2017time}. The contact planner in this framework normally generates kinematically feasible set of contacts \cite{deits2014footstep}. Another approach is to consider the problem of contact planning and momentum trajectory generation simultaneously in a single optimization problem \cite{mordatch2012discovery,winkler2018gait}. Due to the non-convex nature of the centroidal momentum dynamics \cite{orin2013centroidal} and discrete nature of switching contacts, these optimization problems in general boil down to a high dimensional non-convex optimization with combinatorial complexity. As a result, based on the state of the art computational power, these approaches cannot reactively regenerate motion in a Model Predictive Control (MPC) setting. Furthermore, the non-convexity of the problem arises the concern of getting stuck in local minima and also the solutions are very sensitive to the initial guess.

On the contrary to these general approaches, there is a massive amount of effort in the literature to employ model abstraction to regenerate plans fast in face of disturbances \cite{khadiv2016stepping,feng2016robust,khadiv2016step,griffin2017walking}. However, this performance is achieved at the cost of sacrificing generality of motion where these planners cannot deal with general motions with multi-contact phases.

This paper aims at bridging the gap between general-purpose optimization approaches which are high-dimensional and non-convex \cite{mordatch2012discovery,winkler2018gait}, and reactive planners based on abstract model of dynamics which are applicable only to walking on flat terrain \cite{feng2016robust,griffin2017walking,englsberger2015three,khadiv2017robust}. We propose an approach amenable to generate a wide range of motions, while we keep the abstraction and convexity of the problem. This approach plans for various types of motions in three stages with high frequency. The first stage generates the Center of Mass (CoM) vertical trajectory consistent with kinematic constraints of the robot. Then, the second stage generates the whole body motion of the robot consistent with physical constraints. Finally, the third stage optimizes over establishment of an external contact (in addition to feet contact) as well as the exerted force to achieve the task and make the plan feasible. The first two stages are formulated as Quadratic Programs (QP), while the third stage is formulated as a Mixed Integer Quadratic Program (MIQP). 

\begin{figure}[h!]
	\centering
	
\includegraphics[scale=0.8, trim ={5.3cm 12.5cm 5cm 7.3cm},clip]{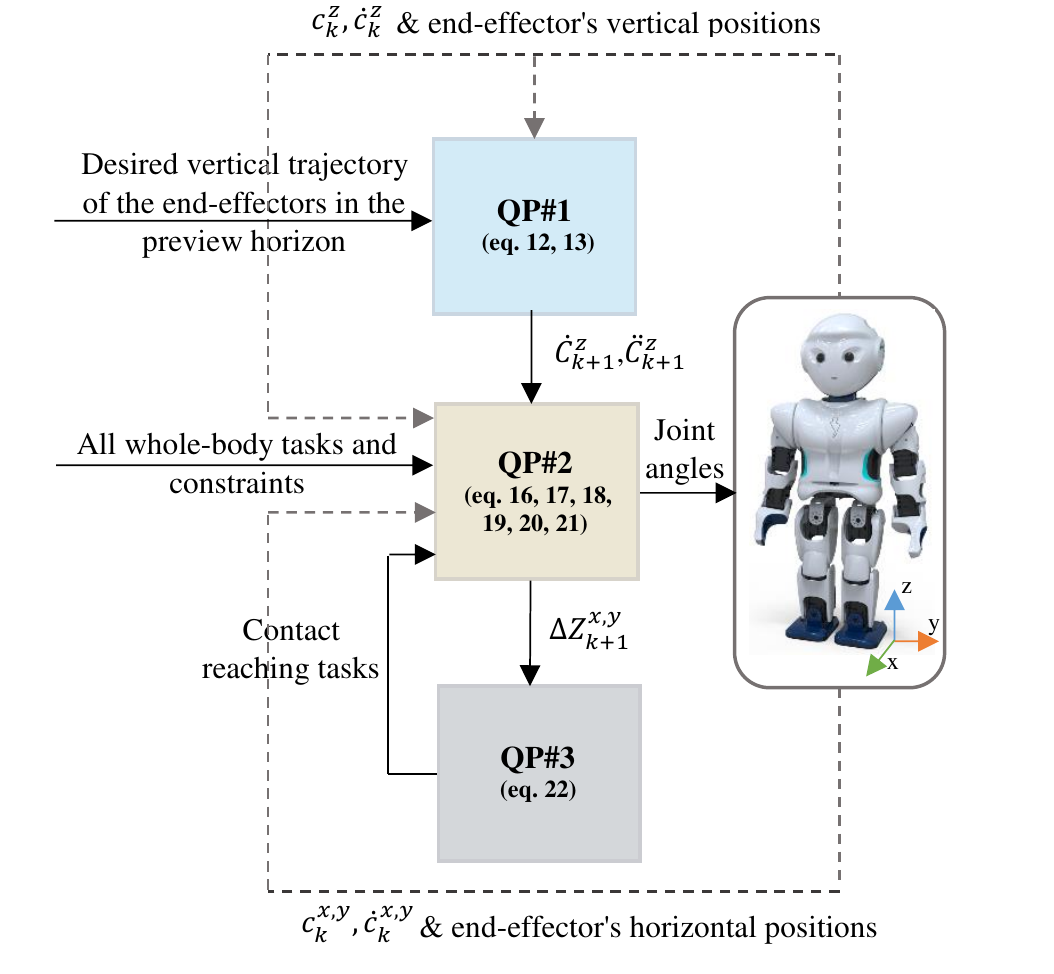}\par
	\caption{Block diagram of the proposed algorithm. The problem is divided into three different stages, each containing a QP designed to plan specific parameters of the motion. CoM height variations, contact forces and other desired parameters can all be planned via quadratic programs due to this separation.}
	\vspace{-0.5em}
	\label{fig1}
\end{figure}

Our proposed algorithm decomposes the whole procedure of pattern generation to three convex optimization problems and enables us to generate whole-body motions in various situations. Some other work used the same strategy for approaching the problem, but in a different fashion. In \cite{mason2018mpc}, the authors proposed a two-stage algorithm that enabled the robot to create or break a contact for maintaining balance when exposed to external disturbances. The paper did not account for CoM height variations or complex whole-body motions. Another important research in the area is \cite{Serra}, in which the authors established an algorithm to generate motions accounting for CoM height variations and external contacts at the same time. However, the choice of contacts were not automatically driven in the algorithm. \cite{AudrenMPC} needs pre-determined contact timings, but it does not make the simplifying assumption of zero angular momentum which can be seen as a good extension for our future work. In general, our approach sums up the advantages of these previous works by introducing a three-stage optimization problem capable of planning both the contacts and the CoM vertical motion automatically. The overview of the approach is briefly illustrated in Fig. \ref{fig1}.

The rest of this paper is as follows: Section II provides the fundamentals required for our optimizer. In section III, the proposed optimization problem is formulated. Section IV summarizes various simulation and experiment scenarios. Finally, Section V concludes the findings.

\section{Fundamentals}

In this section we provide the necessary fundamentals to achieve our goal which is physically feasible whole-body motion planning. To begin, consider the Newton-Euler dynamic equations of a floating-base robotic system with contact:
\begin{equation}
m(\ddot{c}+g)=f_{c}+\sum f_{i}
\label{Newton-Euler1}
\end{equation}
\begin{equation}
c \times m(\ddot{c}+g)+\dot{L}=p \times f_{c}+\sum s_{i} \times f_{i}
\label{Newton-Euler2}
\end{equation}

where $ f_c $ is the external contact force and $ p $ is the corresponding contact location. $ c $ represents the center of mass (CoM) position, $ L $ is the centroidal angular momentum and $ f_{i}$'s are the feet contact forces with locations at $ s_{i} $. Combining the two equations yields (\cite{mason2018mpc}):
\begin{equation}
\dfrac{\sum s_{i} \times f_{i}}{\sum f_{i}^{z}} = \dfrac{c \times m(\ddot{c}+g)-p \times f_{c}+\dot{L}}{m(\ddot{c}^{z}+g)- f_{c}^{z}}
\label{Combining-Nwtn-Eulr}
\end{equation}

Assuming constant angular momentum around the Center of Mass (CoM) $ (\dot{L} = 0) $ and coplanar feet contact, the ZMP equation (on the feet plane) can be obtained as:
\begin{equation}
ZMP^{x,y} = 
\dfrac{m c^{x,y} \ddot{c}^{z} - m c^{z} \ddot{c}^{x,y} + m g c^{x,y} + p^{z} f_{c}^{x,y} - p^{x,y} f_{c}^{z}}
{m(\ddot{c}^{z}+g)- f_{c}^{z}}
\label{ZMP-eq}
\end{equation}

Equation \eqref{ZMP-eq} is the basis of our work in the rest of this paper. Note that under the assumptions of constant CoM height and no external contact forces, \eqref{ZMP-eq} simplifies to the common ZMP formulation for the Linear Inverted Pendulum Model (LIPM):
\begin{equation}
ZMP_{LIPM}^{x,y} = c^{x,y} - \dfrac{c^{z} \ddot{c}^{x,y}}{g}
\label{simplify-ZMPeq}
\end{equation}

Although \eqref{simplify-ZMPeq} is a very simple and useful equation, it can be applied to very limited motion planning problems. In other words, constant CoM height and lacking external contact forces are very restrictive assumptions when planning whole body motions for a humanoid robot. Therefore, in the following we propose an algorithm that overcomes these restrictions while still benefiting from the simplicity of the LIPM.

\section{Approach}
It can be seen from \eqref{ZMP-eq} that the terms corresponding to external contact forces and the CoM vertical acceleration are the ones that make the problem nonlinear. Our approach is to decompose the optimization problem to three different convex QPs, each designed to optimize certain parameters of the motion:\\
\textit{QP $ \# $1: This QP plans the CoM vertical motion consistent with kinematic limitations of the robot.\\
QP $ \# $2: This is the main QP designed to satisfy the whole body tasks and constraints.\\
QP $ \# $3: If the need for an external contact is detected in the previous QP, this QP is fired to plan the contact forces and their corresponding locations.}

Although dividing the optimization problem into three different stages significantly reduces the mathematical complexity, it will result in a sub-optimal solution instead of an optimal one. This will not be an issue since we believe that the computational simplicity and the feasibility of motions are more critical when planning whole-body motions for a humanoid robot.

\subsection{Stage 1: Planning the CoM vertical motion}
We propose to plan the CoM vertical motion in a predictive scheme similar to the approach taken by Model Predictive Control (MPC) to plan the CoM horizontal motions \cite{herdt2010online}. The CoM vertical motion is planned in the preview horizon in order to satisfy the \textit{kinematic limitations}, i. e. the constraints encoding the allowable distance between the CoM of the robot and the desired position of the end-effectors (usually hands or feet). These constraints guarantee the kinematic feasibility of the motion in vertical direction.

For more clarity, consider the case where the robot is given a task to reach an object on the ground. The end-effector of the robot (hand in this case) is constrained to move downwards in a certain direction. In order for the motion to be feasible, the CoM height $ ( c^{z} ) $ must lie within an admissible region compared to the vertical position of hand. This region is determined by the size of the robot and the length of its links. Therefore this kinematic constraint causes the CoM to move in $z$ direction consistent with the desired hand trajectory. Another example is when a robot walks on a set of stairs. In this case, $ c^{z} $ needs to stay in a reasonable area compared to the vertical position of the feet. The proposed QP can successfully plan the CoM vertical motion for both of these cases and many other whole-body scenarios.\\
To formulate the proposed QP, we start with introducing the discrete-time dynamics in $z$ direction:
\begin{equation}
\hat{c}_{k+1}^{z}=
\begin{bmatrix}
1 & T & {{T}^{2}}/2 \\
0 & 1 & T \\
0 & 0 & 1 \\
\end{bmatrix}
\hat{c}_{k}^{z}+
\begin{bmatrix}
{{T}^{3}}/6  \\
{{T}^{2}}/2  \\
T  \\
\end{bmatrix}
\dddot{c}_{k}^{z}
\label{state-eq}
\end{equation}
where:
\begin{equation}
\hat{c}^{z}=
\begin{bmatrix}
{c}^{z}(t_{k}) \\
\dot{c}^{z}(t_{k}) \\
\ddot{c}^{z}(t_{k}) \\
\end{bmatrix},
\dddot{c}_{k}^{z}=\dddot{c}^{z}(kT)
\label{c-hat-z}
\end{equation}

The motion is assumed to have constant jerk $ \dddot{c}^{z} $ over time intervals of length $ T $. Using \eqref{state-eq} and \eqref{c-hat-z} recursively we can derive the equations for CoM vertical position, velocity and acceleration in a preview horizon of length $ NT $:
\begin{equation}
C^{z}_{k+1}=
\begin{bmatrix}
c^{z}_{k+1} \\
\vdots \\
c^{z}_{k+N} \\
\end{bmatrix}
=P_{ps} \hat{c}^{z}_{k} + P_{pu} \dddot{C}^{z}_{k}
\label{N-preview-Cz}
\end{equation}
\begin{equation}
\dot{C}^{z}_{k+1}=
\begin{bmatrix}
\dot{c}^{z}_{k+1} \\
\vdots \\
\dot{c}^{z}_{k+N} \\
\end{bmatrix}
=P_{vs} \hat{c}^{z}_{k} + P_{vu} \dddot{C}^{z}_{k}
\label{N-preview-dpt(Cz)}
\end{equation}
\begin{equation}
\ddot{C}^{z}_{k+1}=
\begin{bmatrix}
\ddot{c}^{z}_{k+1} \\
\vdots \\
\ddot{c}^{z}_{k+N} \\
\end{bmatrix}
=P_{as} \hat{c}^{z}_{k} + P_{au} \dddot{C}^{z}_{k}
\label{N-preview-ddot(Cz)}
\end{equation}

$ P_{ps} $, $ P_{pu} $, $ P_{vs} $ and $ P_{vu} $ are the same as the ones derived in \cite{herdt2010online}. Here we add $ P_{as} $ and $ P_{au} $ as:
\begin{equation}
P_{as}=
{\begin{bmatrix}
0 & 0 & 1 \\
\vdots & \vdots & \vdots \\
0 & 0 & 1 \\
\end{bmatrix}}_{N \times 3}, \quad
P_{au}=
{\begin{bmatrix}
T & 0 & 0 \\
\vdots & \ddots & 0 \\
T & \cdots & T \\
\end{bmatrix}}_{N \times N}
\label{P_as&P_au}
\end{equation}

At each control cycle, the following optimization problem is solved to calculate the CoM vertical trajectory and acceleration in the preview horizon:
\begin{equation}
\begin{aligned}
\min\;
& \norm{\dddot{C}_{k}^{z}}\\
\text{s.t.}\quad &
\abs{{C}^{z}_{k+1} - {R}^{z}_{k+1}} \le d \\
\end{aligned}
\quad,\quad{R}^{z}_{k+1}=
\begin{bmatrix}
\quad r^{z}_{k+1} \\
\vdots \\
r^{z}_{k+N} \\
\end{bmatrix}
\label{QP1}
\end{equation}
where $ r_{k}^{z} $ is the desired position of the end-effector at time instant $ k $ and $ d $ encodes the maximum allowable distance between the COM and the end-effector in $z$ direction. َAdditional constraints can be added to this QP to prevent CoM going higher or lower than certain heights to make sure that the physical limitations of the robot are not violated. This QP yields $ \ddot{C}^{z} $ and $ \dot{C}^{z} $ in the preview horizon which now allows us to proceed to the next QP.
\subsection{Stage 2 : Main QP including whole-body tasks and constraints}
After solving QP$ \# $1 in the previous stage, we are able to plan the whole body motion of the robot consistent with the desired tasks and objectives. With known $ \ddot{C}^{z}  $ and $ \dot{C}^{z} $, the only terms that (if exists) causes nonlinearity in \eqref{ZMP-eq} are the $external$ contact forces (e. g. hand contact). Here we divide the whole body scenarios into two different categories based on whether the external forces and their corresponding locations are known $a priori$ or they need to be planned in the optimization problem. In what follows, we propose two slightly different approaches to deal with each case.

\subsubsection*{a) Scenarios with known force and contact locations}

In this case the external forces along with their corresponding locations are known and the robot needs to manipulate its horizontal CoM jerk to satisfy the balance constraints. Motions that include lifting or moving specific objects fall into this category. In these scenarios, the weight of the object encodes the external force applied to the robot $ ( m_{object} \times g ) $ and the contact locations are specified by the desired position to move the object. With these in mind, \eqref{ZMP-eq} reduces to:
 \begin{equation}
ZMP_{k}^{x,y} = a_{k} c_{k}^{x,y} - b_{k} \ddot{c}_{k}^{x,y} + j_{k}
\label{ZMP_k}
\end{equation}
\begin{equation}
\begin{aligned}
&a_{k}=
\dfrac{mg + m\ddot{c}_{k}^{z}} {{m(\ddot{c}_{k}^{z}+g) - f_{c}^{z}}},\quad
b_{k}=
\dfrac{m c_{k}^{z}} {{m(\ddot{c}_{k}^{z}+g) - f_{c}^{z}}},&\\
&j_{k}=
\dfrac{{p}_{k}^{x,y} - f_{c}^{z}} {{m(\ddot{c}_{k}^{z}+g) - f_{c}^{z}}}&
\end{aligned}
\label{ak_bk_qk}
\end{equation}

where $  p_{k}  $ stands for the position of the external force at time instant $ k $ and $ m $ represents total mass of the robot. Equation \eqref{ZMP_k} is similar to \eqref{simplify-ZMPeq} but with different coefficients. By using \eqref{ZMP_k} recursively we can derive the ZMP vector in the preview horizon:
\begin{equation}
Z^{x,y}_{k+1}=
\begin{bmatrix}
{ZMP}^{x,y}_{k+1} \\
\vdots \\
{ZMP}^{x,y}_{k+N} \\
\end{bmatrix}
=P_{zs} \hat{c}^{x,y}_{k} + P_{zu} \dddot{C}^{x,y}_{k} + P_{zj}
\label{ZMP_k+1}
\end{equation}
where:
\begin{equation}
\begin{aligned}
&P_{zs}=
{\begin{bmatrix}
	a_{k+1} & T a_{k+1} & \dfrac{T^2}{2} a_{k+1} - b_{k+1} \\
	\vdots & \vdots & \vdots \\
	a_{k+N} & NT a_{k+N} & \dfrac{N^2 T^2}{2} a_{k+N} -b_{k+N} \\
	\end{bmatrix}}, \quad P_{zu}= &\\
&{\begin{bmatrix}
	\dfrac{T^3}{6} a_{k+1} - T b_{k+1} & 0 & 0 \\
	\vdots & \ddots & 0 \\
	(1-3N+3N^2) \dfrac{T^3}{6} a_{k+N} - T b_{k+N} & \cdots & \dfrac{T^3}{6} a_{k+1} - T b_{k+1} \\
	\end{bmatrix}},&\\
&P_{zj}=
\begin{bmatrix}
{j}_{k+1} \\
\vdots \\
{j}_{k+N} \\
\end{bmatrix}&
\end{aligned}
\label{Pz-eqs}
\end{equation}

The ZMP formulation derived in \eqref{ZMP_k+1} will be used to   satisfy the ZMP constraint:
\begin{equation}
Z^{x,y}_{low} \le Z^{x,y}_{k+1} \le Z^{x,y}_{up}
\label{ZMP-constraint}
\end{equation}

It is noteworthy that this ZMP bound is defined for the case that the feet are co-planar. In the case that the feet are not co-plannar, we can define the ZMP and its bounds on an arbitrary plane using the approach in \cite{caron2017zmp}.\\
This balance constraint together with other whole-body tasks and objectives (see \cite{sherikov2014whole}) can now be solved in a QP to plan a stable whole body motion for the robot.\\

\subsubsection*{b) Scenarios with unknown force and contact locations}

The majority of whole body scenarios fall into this category where the robot has to decide whether it needs to make a contact or not; and if it does, the choice of contact and the contact forces must be planned ahead. However, we can see from \eqref{ZMP-eq} that the external force $ f_{z} $ makes the formulation nonlinear which renders the corresponding optimization problem non-convex. \cite{mason2018mpc} proposed an algorithm to deal with this non-linearity and used it for balance recovery scenarios. Here we extend this method to plan various whole-body motions such as reaching an object or passing through a hole.

To make the ZMP constraint linear we introduce the parameter $ \Delta Z $ which presents the difference between the ZMP caused by hand contacts $ ( ZMP_{contact}^{x,y} ) $ and the simple LIPM case $ ( ZMP_{LIPM}^{x,y} ) $:
\begin{equation}
\begin{aligned}
&\ Z^{x,y}_{contact} = \ Z^{x,y}_{LIPM} + \Delta Z^{x,y}&\\
&\Delta Z^{x,y}=
\begin{bmatrix}
{\Delta Z}^{x,y}_{k+1} \\
\vdots \\
{\Delta Z}^{x,y}_{k+N} \\
\end{bmatrix}&
\end{aligned}
\label{DeltaZ}
\end{equation}
\begin{equation}
\begin{aligned}
&\Delta Z^{x,y}_{k} = 
(\dfrac{{p}_{k}^{z}} {{m(\ddot{c}_{k}^{z}+g) - f_{c}^{z}}}) f_{c}^{x,y}+&\\
&(\dfrac{ - p_{k}^{x,y} - \dfrac{c_{k}^{z} {\ddot{c}}_{k}^{x,y}}{g} + c_{k}^{x,y}} {{m(\ddot{c}_{k}^{z}+g) - f_{c}^{z}}}) f_{c}^{z}+
\dfrac{m c_{k}^{z} {\ddot{c}}_{k}^{x,y} {\ddot{c}}_{k}^{z}} {{mg(\ddot{c}_{k}^{z}+g) - g f_{c}^{z}}}&\\
\end{aligned}
\label{DeltaZ_k}
\end{equation}

Equations \eqref{DeltaZ} and \eqref{DeltaZ_k} can now be used to impose ZMP constraints in the preview horizon:
\begin{equation}
Z^{x,y}_{low} \le {ZMP}^{x,y}_{LIPM} + {\Delta Z}^{x,y} \le Z^{x,y}_{up}
\label{ZMP-constraints-in-N-horizon}
\end{equation}

By adding the parameter $ {\Delta Z}^{x,y} $ to the decision variables of our optimization problem, we can postulate if there is any need for a contact: As long as all the $ {\Delta Z}^{x,y} $ values in the preview horizon are equal or close to zero, the robot is able to maintain its balance without the help of a contact. However, once QP$\# $2 calculates a non-zero value for $ {\Delta Z}^{x,y} $ anywhere in the preview horizon, QP$\# $3 is fired to choose the proper contact among all the available contact locations and to estimate the contact forces \cite{sherikov2014whole}. For more clarification, consider a scenario in which the robot is supposed to reach an object. The basic QP for this case would be:
\begin{equation}
\begin{aligned}
\underset{ \dddot{C}^{x,y}, \Delta{Z^{x,y}} , \dot{q} }{\min} \
& {\alpha}_{1} \norm{\dddot{C}^{x,y}} + {\alpha}_{2} \norm{\Delta{Z}^{x,y}} + {\alpha}_{3} \norm{\dot{q}} + \\
&{\alpha}_{4} \norm{J_{CoM}^{x,y,z} \dot{q} - {\dot{c}}^{x,y,z}} + {\alpha}_{5} \norm{J_{hand}^{x,y,z} \dot{q} - {V}_{hand}^{x,y,z}} \\
\text{s.t.} \quad &
Eq.\;\ref{ZMP-constraints-in-N-horizon} \quad and \quad \abs{R_{k+1}^{x,y} - C_{k+1}^{x,y}} \le{d}
\end{aligned}
\label{formul}
\end{equation}
where $ \ J_{CoM} $ and $ \ J_{hand} $ represent Jacobian matrices for CoM and hand positions respectively.  $ \dot{q} $ stands for joint velocities and $ \ V_{hand} $ indicates the desired velocity for the hand end-effector. An inequality constraint is added to account for the maximum allowable distances between CoM and hands in $ x $ and $ y $ directions. Additional constraints can be considered to determine the position of feet or to add any other desired tasks \cite{sherikov2014whole}. 

\subsection{Stage 3: planning the contact location and forces}
In this stage a Mixed Integer Quadratic Program (MIQP) is employed to choose the optimal contact location and to calculate the contact forces \cite{mason2018mpc}. We can rearrange \eqref{DeltaZ_k} to:
\begin{equation}
\begin{aligned}
&f^{x,y}_{c} = 
(\dfrac{{p}^{x,y} + \dfrac{c^z {\ddot{c}}^{x,y}}{g} - c^{x,y} -{\Delta Z}^{x,y} } {p^z}) f_{c}^{z}+&\\
&+\dfrac{m (g + {\ddot{c}}^z) {\Delta Z}^{x,y} - \dfrac{m c^z {\ddot{c}}^{x,y}}{g} {\ddot{c}}^z} {p^z}&\\
\end{aligned}
\label{neweq}
\end{equation}

It can be seen from \eqref{DeltaZ_k} that $ f_{c}^{x,y} $ is a linear function of $ f_{c}^{z} $. As a result, we only need to solve the optimization problem for $ f_{c}^{z} $. Also to avoid slippage, linearized friction cone constraints are added \cite{khadiv2017pattern}.


To choose between the available contacts, we introduce a binary variable $ H_{i} $ for each contact $ i $. Big-M formulation method \cite{lofberg2012big} is exploited to include the binary variables in our QP constraints. Hence, the optimization problem will be of the form:
\begin{equation}
\begin{aligned}
\underset{{{f}_{c}^{z}}, H_1 , H_2 , \ldots }{\min} \
& \norm{f_{c}^{z}} + \norm{w_1 H_1} + \norm{w_2 H_2} + \ldots\\
\text{s.t.} \,\,\, &
Friction \,\,\, Cone \,\,\, Constraints , \,\,\,   H_1+H_2+...=1, \,\,\,  \\ 
\end{aligned}
\label{QP2}
\end{equation}

\begin{figure*}
	\centering
	\includegraphics[scale=1.01, trim ={1.7cm 22cm 1.1cm 2cm},clip]{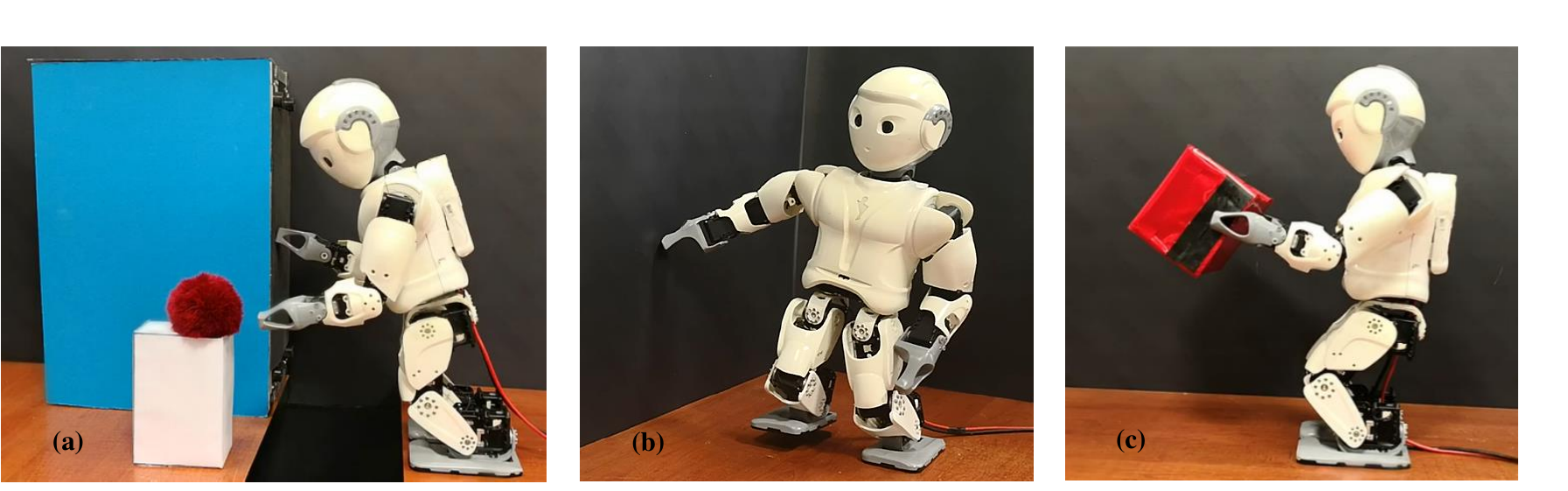}\par
	\caption[Caption for LOF]{Multi-contact experiments implemented on SURENA-MINI humanoid robot using the proposed algorithm}
	\vspace{-1.5em}
	\label{fig2}
\end{figure*}

Weighting coefficients $ w_{i} $ are used to give priority to closer contacts. Also, it is assumed that at least one of the available contacts is located at a reachable area with respect to the robot. This assumption is necessary to make sure that reaching a contact would not require any changes in the CoM height, as the CoM vertical motion is already planned in QP$ \# $1. This assumption seems legitimate since the contacts are supposed to \textit{help} the robot to achieve a more complicated goal. The above MIQP can be solved using the GUROBI optimization software \cite{optimization2014inc}.

\section{Simulations and Experiments}
In this section, several simulations and experiments are demonstrated to show the validity of the proposed algorithm. All the experiments are implemented on SURENA-MINI humanoid robot \cite{NikkhahMini}. The accompanying video contains all the presented simulations and experiments as it is also briefly demonstrated in Fig. 2\footnote{The video is available on https://youtu.be/RQr1gmuosSQ .} .

We start with the case where the robot is supposed to reach an object outside of its reachable area (Fig. \ref{fig2}.a). In this case, since the height of the object is neither too high nor too low, there is no need to change the CoM height in order to satisfy the constraints of the first QP. In the second QP, similar kinematic constraints are checked for $x$ and $y$ directions of the CoM. Here the position of the object in $x$ direction is out of the reachable area. Therefore the kinematic constraints of the second QP imply that the CoM needs to \textit{follow} the hand in order for the motion to be feasible. With the CoM moving forward, the second optimization may detect a need for an external help to satisfy the ZMP constraint. As soon as this happens, the third QP is fired to select a new contact point. This new contact will then be added to QP $ \# $2 in the next iteration so that the free hand (right hand in this case) will reach it in the planned time (see Fig. \ref{fig1}).

\begin{figure}[h!]
	\centering
	\includegraphics[scale=0.9, trim ={5.7cm 8.8cm 4.5cm 10cm},clip]{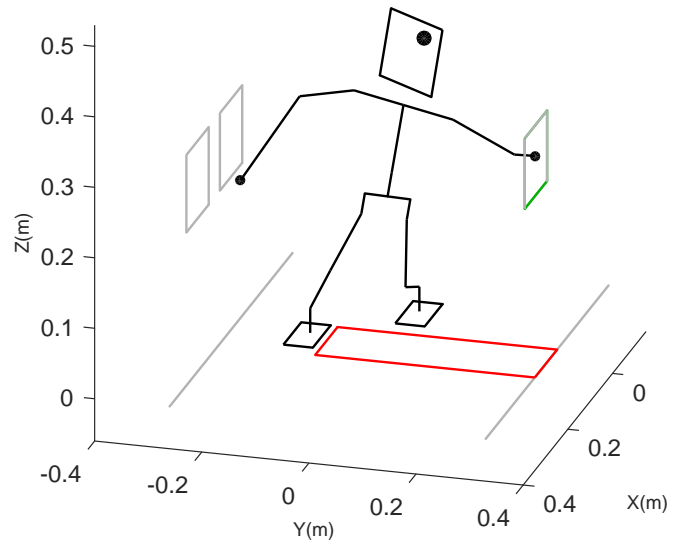}\par
	\caption{Multi-contact walking for a humanoid robot planned by the proposed algorithm. The choice of contact location, the time to create or break a contact and the corresponding contact forces are all decided automatically within the algorithm.}
	\vspace{-1.5em}
	\label{fig3}
\end{figure}

In our second scenario, the robot is required to make a large step in $y$ direction (Fig. \ref{fig2}.b). The generality of the proposed algorithm lets us to treat this scenario similar to the previous one: The CoM has to follow the stance foot in $y$ direction to satisfy the kinematic constraints in the second QP. These constraints are imposed due to length of the robots legs and the limitation on the hip joint roll angle. With the CoM moving in $y$ direction, it would not be possible for the robot to maintain its balance without the help of an external contact. As a result, the third QP is fired to choose the contact location and its corresponding contact forces. This scenario can be easily extended for more practical uses. As shown in Fig. \ref{fig3} the robot is supposed to pass through a hole (the red rectangle). The presented scheme plans a sequence of contacts for this scenario where the suitable times to create or break a contact are all computed automatically according to the elements in $ \Delta Z $ vector.

\begin{figure}[h!]
	\centering
	\includegraphics[scale=0.91, trim ={6.5cm 9.7cm 4.5cm 10cm},clip]{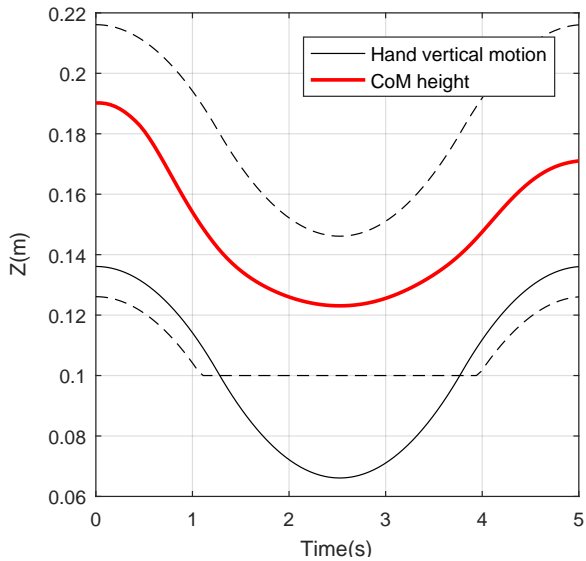}\par
	\caption{The planned trajectory for CoM height in the box lifting scenario. The dashed lines represent the upper and lower bounds of the admissible region for CoM.}
	\vspace{-1.5em}
	\label{fig4}
\end{figure}

The previously discussed experiments did not need any significant changes in the CoM height. But in real-life situations there are lots of scenarios in which the robot needs to adjust its CoM vertical position in order to perform a given task. Therefore for the third experiment, we consider a task where the robot has to pick a box from the floor and lift it up to a certain height. In this case, the height of the CoM needs to change so that the generated motion is consistent with the physical limitations of the robot. This is considered by the first QP which implies that the location of the box is farther than it can be reached by the hand movements alone. Therefore, the motion of the CoM in $z$ direction is adapted to follow the hands. Fig.\ref{fig4} shows the planned trajectory for CoM vertical motion along with the trajectory of hands in $z$ direction. The dashed lines represent the upper and lower bounds of the admissible region for the CoM height. These bounds are created by the maximum allowable distance between the CoM and the hand positions. Note that at some instances the lower bound of the CoM height changes to a straight line. This is caused by the fact that the CoM can not get lower than a certain height due to joint limits. It can be seen from Fig.\ref{fig4} that the algorithm has successfully planned a smooth trajectory for the CoM height within the admissible region. Also, for more robustness, a weakly weighted objective can be added to maintain the CoM height in the middle of the admissible region. This simulation can be easily extended to the case where the robot needs to carry a box while walking on an uneven train or on a set of stairs.

\section{CONCLUSION} 
In this paper, we presented a multi-contact motion generation algorithm while planning the required changes for the CoM height. The presented method uses the model predictive control scheme and decomposes the problem to three different stages. This decomposition lets us treat the whole nonlinear problem as a set of three quadratic programs that can be solved efficiently at each control cycle. This results in a scheme that is simple and suitable for real-time implementations and also is flexible enough to plan a vast majority of whole-body motions. The addition of two stages to the algorithm, each containing a quadratic program, makes it possible to overcome the common simplifying assumptions of linear inverted pendulum model, i. e. constant CoM height and no external contact forces. The validity of the proposed algorithm was demonstrated through several simulations and experiments.

%
%
%

\bibliography{Master}

\begin{thebibliography}{10}

\bibitem{darpa2015drc}
DARPA.
\newblock Darpa robotics challenge, 2015.

\bibitem{dai2016planning}
Hongkai Dai and Russ Tedrake.
\newblock Planning robust walking motion on uneven terrain via convex
  optimization.
\newblock In {\em Humanoid Robots (Humanoids), 2016 IEEE-RAS 16th International
  Conference on}, pages 579--586. IEEE, 2016.

\bibitem{carpentier2017multi}
Justin Carpentier and Nicolas Mansard.
\newblock Multi-contact locomotion of legged robots.
\newblock {\em submitted to IEEE Transactions on Robotics}, 2017.

\bibitem{ponton2017time}
Brahayam Ponton, Alexander Herzog, Stefan Schaal, and Ludovic Righetti.
\newblock On time optimisation of centroidal momentum dynamics.
\newblock In {\em (ICRA), 2018 IEEE International Conference on Robotics and
  Automation}, pages 1--7. IEEE, 2018.

\bibitem{deits2014footstep}
Robin Deits and Russ Tedrake.
\newblock Footstep planning on uneven terrain with mixed-integer convex
  optimization.
\newblock In {\em Humanoid Robots (Humanoids), 2014 14th IEEE-RAS International
  Conference on}, pages 279--286. IEEE, 2014.

\bibitem{mordatch2012discovery}
Igor Mordatch, Emanuel Todorov, and Zoran Popovi{\'c}.
\newblock Discovery of complex behaviors through contact-invariant
  optimization.
\newblock {\em ACM Transactions on Graphics (TOG)}, 31(4):43, 2012.

\bibitem{winkler2018gait}
Alexander~W Winkler, C~Dario Bellicoso, Marco Hutter, and Jonas Buchli.
\newblock Gait and trajectory optimization for legged systems through
  phase-based end-effector parameterization.
\newblock {\em IEEE Robotics and Automation Letters}, 3(3):1560--1567, 2018.

\bibitem{orin2013centroidal}
David~E Orin, Ambarish Goswami, and Sung-Hee Lee.
\newblock Centroidal dynamics of a humanoid robot.
\newblock {\em Autonomous Robots}, 35(2-3):161--176, 2013.

\bibitem{khadiv2016stepping}
Majid Khadiv, Sebastien Kleff, Alexander Herzog, S.~Ali~A. Moosavian, Stefan
  Schaal, and Ludovic Righetti.
\newblock Stepping stabilization using a combination of dcm tracking and step
  adjustment.
\newblock In {\em Robotics and Mechatronics (ICRoM), 2016 4th RSI International
  Conference on}, 2016.

\bibitem{feng2016robust}
Siyuan Feng, X~Xinjilefu, Christopher~G Atkeson, and Joohyung Kim.
\newblock Robust dynamic walking using online foot step optimization.
\newblock In {\em Intelligent Robots and Systems (IROS), 2016 IEEE/RSJ
  International Conference on}, pages 5373--5378. IEEE, 2016.

\bibitem{khadiv2016step}
Majid Khadiv, Alexander Herzog, S~Ali~A Moosavian, and Ludovic Righetti.
\newblock Step timing adjustment: A step toward generating robust gaits.
\newblock In {\em Humanoid Robots (Humanoids), 2016 IEEE-RAS 16th International
  Conference on}, pages 35--42. IEEE, 2016.

\bibitem{griffin2017walking}
Robert~J Griffin, Georg Wiedebach, Sylvain Bertrand, Alexander Leonessa, and
  Jerry Pratt.
\newblock Walking stabilization using step timing and location adjustment on
  the humanoid robot, atlas.
\newblock {\em arXiv preprint arXiv:1703.00477}, 2017.

\bibitem{englsberger2015three}
Johannes Englsberger, Christian Ott, and Alin Albu-Sch{\"a}ffer.
\newblock Three-dimensional bipedal walking control based on divergent
  component of motion.
\newblock {\em IEEE Transactions on Robotics}, 31(2):355--368, 2015.

\bibitem{khadiv2017robust}
Majid Khadiv, Alexander Herzog, S.~Moosavian, and Ludovic Righetti.
\newblock A robust walking controller based on online step location and
  duration optimization for bipedal locomotion.
\newblock {\em arXiv preprint arXiv:1704.01271}, 2017.

\bibitem{mason2018mpc}
Sean Mason, Nicholas Rotella, Stefan Schaal, and Ludovic Righetti.
\newblock An mpc walking framework with external contact forces.
\newblock In {\em 2018 IEEE International Conference on Robotics and Automation
  (ICRA)}, pages 1785--1790. IEEE, 2018.

\bibitem{Serra}
Diana Serra, Camille Brasseur, Alexander Sherikov, Dimitar Dimitrov, and
  Pierre-Brice Wieber.
\newblock A newton method with always feasible iterates for nonlinear model
  predictive control of walking in a multi-contact situation.
\newblock In {\em Humanoid Robots (Humanoids), 2016 IEEE-RAS 16th International
  Conference on}, pages 932--937. IEEE, 2016.

\bibitem{AudrenMPC}
Herve Audren, Joris Vaillant, ne~Kheddar, Abderrahma, Adrien Escande, Kenji
  Kaneko, and Eiichi Yoshida.
\newblock Model preview control in multi-contact motion–application to a
  humanoid robot.
\newblock In {\em Intelligent Robots and Systems (IROS), 2014 IEEE/RSJ
  International Conference on}, pages 4030--4035. IEEE, 2014.

\bibitem{herdt2010online}
Andrei Herdt, Holger Diedam, Pierre-Brice Wieber, Dimitar Dimitrov, Katja
  Mombaur, and Moritz Diehl.
\newblock Online walking motion generation with automatic footstep placement.
\newblock {\em Advanced Robotics}, 24(5-6):719--737, 2010.

\bibitem{caron2017zmp}
St{\'e}phane Caron, Quang-Cuong Pham, and Yoshihiko Nakamura.
\newblock Zmp support areas for multicontact mobility under frictional
  constraints.
\newblock {\em IEEE Transactions on Robotics}, 33(1):67--80, 2017.

\bibitem{sherikov2014whole}
Alexander Sherikov, Dimitar Dimitrov, and Pierre-Brice Wieber.
\newblock Whole body motion controller with long-term balance constraints.
\newblock In {\em Humanoid Robots (Humanoids), 2014 14th IEEE-RAS International
  Conference on}, pages 444--450. IEEE, 2014.

\bibitem{khadiv2017pattern}
Majid Khadiv, S~Ali~A Moosavian, Alexander Herzog, and Ludovic Righetti.
\newblock Pattern generation for walking on slippery terrains.
\newblock In {\em Robotics and Mechatronics (ICRoM), 2017 5th RSI International
  Conference on}, 2017.

\bibitem{lofberg2012big}
Johan Löfberg.
\newblock Big-m and convex hulls, 2012.

\bibitem{optimization2014inc}
Gurobi~Optimization Inc.
\newblock Gurobi optimizer reference manual, 2014.

\bibitem{NikkhahMini}
Arman Nikkhah, Aghil Yousefi-Koma, Reihaneh Mirjalili, and Hossein
  Morvaridi~Farimani.
\newblock Design and implementation of small-sized 3d printed surena-mini
  humanoid platform.
\newblock In {\em Robotics and Mechatronics (ICRoM), 2017 5th RSI International
  Conference on}, 2017.

\end{thebibliography}
\bibliographystyle{unsrt}

\end{document}